\documentclass{article}

\usepackage{microtype}
\usepackage{graphicx}
\usepackage{caption}
\usepackage{subcaption}
\usepackage{booktabs} 

\usepackage{hyperref}

\usepackage[accepted]{icml2024}

\usepackage{amsmath}
\usepackage{amssymb}
\usepackage{mathtools}
\usepackage{amsthm}

\usepackage[capitalize,noabbrev]{cleveref}

\theoremstyle{plain}

\theoremstyle{definition}

\theoremstyle{remark}

\usepackage[textsize=tiny]{todonotes}

\newcommand{\bs}[1]{\boldsymbol{#1}}
\newcommand{\ie}{i.e.\,}
\newcommand{\eg}{e.g.\,}

\icmltitlerunning{
Tilting the Odds at the Lottery
}
\allowdisplaybreaks

\begin{document}

\twocolumn[
\icmltitle{
Tilting the Odds at the Lottery:\\the Interplay of Overparameterisation and Curricula in Neural Networks
}




\begin{icmlauthorlist}
\icmlauthor{Stefano Sarao Mannelli}{ucl}
\icmlauthor{Yaraslau Ivashynka}{bocconi}
\icmlauthor{Andrew Saxe}{ucl}
\icmlauthor{Luca Saglietti}{bocconi}
\end{icmlauthorlist}

\icmlaffiliation{ucl}{Gatsby Computational Neuroscience Unit \& Sainsbury Wellcome Centre, University College London, London, UK}
\icmlaffiliation{bocconi}{Department of Computing Sciences, Bocconi University, Milano, Italy}

\icmlcorrespondingauthor{Stefano Sarao Mannelli}{s.saraomannelli@ucl.ac.uk}
\icmlcorrespondingauthor{Luca Saglietti}{luca.saglietti@unibocconi.it}

\icmlkeywords{Machine Learning, ICML}

\vskip 0.3in
]



\printAffiliationsAndNotice{} 

\begin{abstract}
A wide range of empirical and theoretical works have shown that overparameterisation can amplify the performance of neural networks. 
According to the \emph{lottery ticket hypothesis}, overparameterised networks have an increased chance of containing a sub-network that is well-initialised to solve the task at hand.
A more parsimonious approach,
inspired by animal learning, consists in guiding the learner towards solving the task by curating the order of the examples, \ie providing a \textit{curriculum}. However, this learning strategy seems to be hardly beneficial in deep learning applications.
In this work, we undertake an analytical study that connects curriculum learning and overparameterisation. In particular, we investigate their interplay in the online learning setting for a 2-layer network in the XOR-like Gaussian Mixture problem.
Our results show that a high degree of overparameterisation---while simplifying the problem---can limit the benefit from curricula, providing a theoretical account of the ineffectiveness of curricula in deep learning.
\looseness=-1
\end{abstract}

\section{Introduction}

The ineffectiveness of curriculum learning in training deep networks is a puzzling empirical observation.
Simple experiments \cite{bengio2009curriculum} and recent theoretical results \cite{weinshall2018curriculum,abbe2021staircase,saglietti2022analytical,sorscher2022beyond,cornacchia2023mathematical,tong2023adaptive} point out several potential benefits from adopting this learning paradigm in a machine learning context, including learning speed-ups and better asymptotic generalisation. However, curriculum learning has not entered the standard pipeline of deep learning applications, and large-scale studies show that indeed for these models the curriculum benefits seem to be marginal \cite{wu2020curricula}.

The apparent contradiction between theory and applications is at the centre of this study. We address this issue by identifying a potential cause: the high degree of overparameterisation, an ever-present feature of deep learning, which can erode the margin of improvement of curriculum learning. In particular, we provide an extensive analytical study of curriculum learning and over-parameterisation in the XOR-like Gaussian mixture model \cite{refinetti2021classifying, arous2023high}. As we will see, this learning problem encapsulates a crucial ingredient of hard learning problems ---the coexistence of a few relevant features hidden among many irrelevant ones. From our study, we obtain the following \emph{main contributions}:
\begin{itemize}
    \item We disentangle the advantages of curriculum learning, giving a detailed account of its impact on \textit{relevant manifold discovery} and \textit{labelling rule identification}, and we show that curriculum can compensate for a wide range of poor initialisations;
    \item We connect these results with the observation that---according to the lottery ticket hypothesis \cite{frankle2018lottery}---a similar role can be played by overparameterisation. 
    \item We investigate whether overparameterisation and curriculum can create a positive synergy in boosting learning performance, and find that eventually---if the degree of overparameterisation is sufficiently high---the effect of curricula vanishes;
    \item Finally, we test our predictions in realistic datasets, empirically confirming the results.
\end{itemize}

\paragraph{Related works.} The theory of curriculum learning in ML is still in its early stages of development. In \citet{weinshall2018curriculum,weinshall2020theory}, the authors analysed the effects on convergence speed of presenting easy vs. hard samples, showing that easy samples lead to faster learning. However, these studies did not account for the importance of ordering within a dataset. In \citet{abbe2021staircase}, in the context of a function approximation task with neural networks, it was shown that when the target function is a high-degree polynomial, guiding the neural networks using lower-degree cues can drastically speed up learning. Similarly, \citet{abbe2023provable, cornacchia2023mathematical} showed that presenting samples non-uniformly in a parity learning task can also lead to a remarkable speed-up. 
The observation of a dynamical effect without strong asymptotic generalisation benefits was also found in a convex learning setting in \citet{saglietti2022analytical}, where the authors analysed a simple model of curriculum proposed in \citet{bengio2009curriculum}. To improve the generalisation gain, the authors proposed the introduction of an \emph{ad hoc} memory term in the loss function.

In the present work, we move beyond convex learning settings and consider a more complex task-learner pair, where a curriculum protocol can in principle achieve a large performance gap at long times without loss alterations. Here, we introduce the crucial ingredient of overparameterisation into the mix, and analyse its interaction with the curriculum strategy.

\section{XOR-like Gaussian Mixture and the effect of over-parametrisation} 

\begin{figure*}[h]
    \vskip 0.2in
    \begin{center}
        \begin{subfigure}[b]{0.29\textwidth}
            \includegraphics[width=\textwidth]{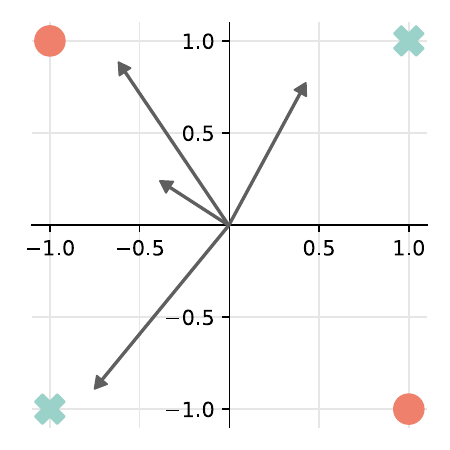}
            \vspace{-0.2cm}
            \caption{Minimal parameterisation}
        \end{subfigure}
        \hfill
        \begin{subfigure}[b]{0.29\textwidth}
            \includegraphics[width=\textwidth]{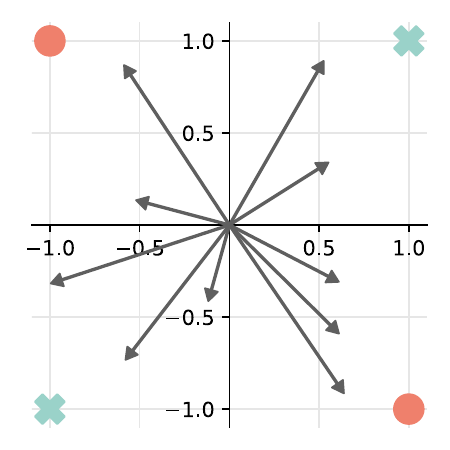}
            \vspace{-0.2cm}
            \caption{Overparameterised NN}
        \end{subfigure}
        \hfill
        \begin{subfigure}[b]{0.4\textwidth}
            \includegraphics[width=\textwidth]{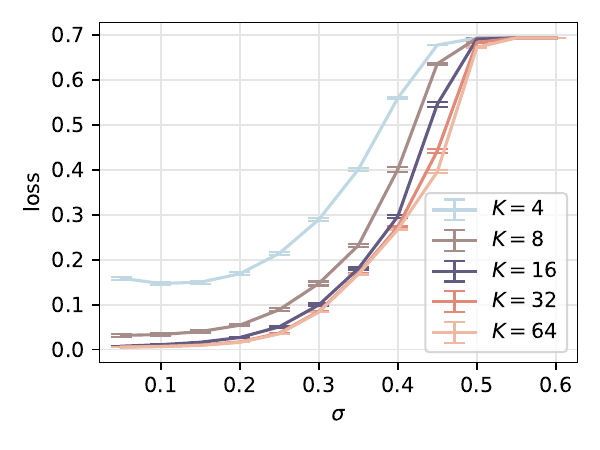}
            \caption{The effect overparameterisation}
        \end{subfigure}
    \end{center}
    \caption{
    \textbf{XOR-like Gaussian mixture and the lottery ticket.} 
    \textit{Panels (a)} and \textit{(b)} show the relevant coordinates in the input space of the Gaussian mixture considered in our theoretical framework. Circles and crosses represent the position of the centroids for the two classes. All the remaining dimensions are spurious, and their centroids are centered at zero. \textit{Panel (a)} represents a possible `bad' initialisation of a 2-layer neural network with $K=4$ hidden units (weight vectors depicted in gray), where `bad' means that this will not lead to the optimal solution. 
    Notice that $K=4$ is however the minimal number of parameters to optimally solve the task.
    \textit{Panel (b)} considers instead an overparameterised neural network with $K=10$: notice that by increasing $K$ the probability of covering all 4 quadrants at initialisation increases, making it more likely to have a good `lottery ticket'. 
    \textit{Panel (c)} shows that, indeed, overparameterisation consistently helps achieve a lower generalisation error (y-axis), even as noise intensity $\sigma$ increases. Eventually, the clusters become so overlapping that even an overparameterised system cannot achieve good performance.  
    }
    \label{fig:fig1}
\end{figure*}

We revisit the XOR-like Gaussian Mixture (XGM) setting previously analysed in \citet{refinetti2021classifying, arous2023high}, which involves a combination of high-dimensional and low-dimensional learning tasks. In particular, an XOR classification problem---see sketches in Fig.~\ref{fig:fig1}---is hidden in high-dimension and is to be discovered and solved by the learning model. Because of the non-linearly separable nature of the XOR problem, at least a hidden layer is required for a neural network to be able to achieve optimal generalisation in this setting. 

\textbf{Model definition.} Given two orthogonal unit vectors $\boldsymbol{\mu}_1,\boldsymbol{\mu}_2 \in \mathbb{R}^d$, we obtain i.i.d.~samples $(\bs X,y)$, with $\bs X\in\mathbb{R}^d$ and $y\in\{0,1\}$, from the following Markov chain: first the label is obtained as $y\sim\mathrm{Ber}(\frac{1}{2})$, then if $y=1$ the inputs follow the mixture $\bs X \sim \frac{1}{2} \mathcal{N}(\bs{\mu}_1, \sigma^2 \mathbb{I}) + \frac{1}{2} \mathcal{N}(-\bs{\mu}_1, \sigma^2 \mathbb{I})$, while if $y=0$ the input distribution is $\bs X\sim \frac{1}{2} \mathcal{N}(\bs{\mu}_2, \sigma^2 \mathbb{I}) + \frac{1}{2} \mathcal{N}(-\bs{\mu}_2, \sigma^2 \mathbb{I})$.
The standard deviation $\sigma$ controls the signal to noise ratio of the problem $\mathrm{SNR} = \frac{\lVert \bs \mu \rVert}{\sigma} = \frac{1}{\sigma}$. Hereafter, we will use the short-hand notation $X^{\pm\alpha}$ for indicating a sample from cluster $\mathcal{N}(\pm \bs\mu_\alpha, \sigma \mathbb{I})$. 

We consider a 2-layer neural network with $K$ hidden units, with parameters $\bs W \in \mathbb{R}^{K\times d}$, $\bs b \in \mathbb{R}^K$, $\bs v \in \mathbb{R}^K$ and activation:
\begin{equation}
    \hat y = \frac1{\sqrt{K}} \sum_{k=1}^{K} v_k g \left(\bs{W}_k \cdot \bs X + b_k\right).
\end{equation}
We assume the first layer weights are initialised according to the typical scaling $W_{ki} \sim \mathcal{N}(0, \frac{1}{\sqrt{d}})$, while the second layer weights are standard Gaussian $v_k\sim\mathcal{N}(0,1)$ and the bias terms are initially set to $b_k=0$. While the non-linearity $g$ could be arbitrary, in the following we will consider the case of ReLU activation, and discuss logistic regression with cross-entropy loss $L(y,\hat{y}) = -y \hat{y} + \log\left( 1 + \exp(\hat{y})\right)$. We focus on the online learning setting (ballistic limit in \citet{arous2023high}) where at each epoch a new pattern is presented and a step of gradient descent is applied to the parameters of the network.

\textbf{ODE limit.} As in \citet{refinetti2021classifying, arous2023high}, we consider the high-dimensional limit where the input dimension $d\to\infty$ and the gradient descent step is rescaled with $1/d$. In this regime, the learning dynamics provably approaches a deterministic limit that can be fully captured by a system of ODEs, which track the evolution of a set of key order parameters. 
In particular, given a sample $X^{\pm \alpha}$, the scalar fields $\{\lambda_i\}_1^{K+2}$ defined as:
\begin{equation}
\lambda_k = \bs W_k \cdot \bs X^{\pm \alpha}; \,\,\,\,\,\lambda_{K+\beta} = \bs \mu_\beta \cdot \bs X^{\pm \alpha}; 
\end{equation}
with $k \in \lfloor K \rfloor $ and $\beta\in\lfloor 2\rfloor$, can be shown to follow a multivariate Gaussian distribution with mean and covariance:
\begin{equation}
    m = [ \pm M_{\cdot \alpha}, T_{\cdot\alpha} ]; \,\,\,\,\,\,C = \sigma^2 \begin{bmatrix}Q & M \\ M^\top & T \end{bmatrix};
\end{equation}
parameterised by the order parameters
\begin{equation*}
Q = \bs W \bs W^\top, \,\,\,\, M_{\cdot \alpha} = \bs W \bs \mu_\alpha, \,\,\,\, T_{\alpha \beta} = \bs\mu_\alpha \cdot \bs\mu_\beta,
\end{equation*}
which respectively represent the overlap matrix between the neurons, the alignment of the neurons with the centroids, and the overlap between the centroids. Note that in our setting $T = \mathbb{I}^{2\times2}$. 

Then, denoting with $\sigma(\cdot)$ the sigmoid function and using the short-hand notation $\Delta=y-\sigma\left(\frac{1}{\sqrt{K}} \sum_k v_k g(\lambda_k + b_k)\right)$, we can define the expectations:
\begin{align*}
A_{ij} &= \mathbb{E}_{\pm\alpha} \mathbb{E}_{\bs{\lambda}} \left[ \lambda_i g'(\lambda_j + b_j) \Delta \right]; \\
B_{ij} &= \sigma^2 \mathbb{E}_{\pm\alpha} \mathbb{E}_{\bs{\lambda}} \left[ g'(\lambda_i + b_i) g'(\lambda_j + b_j) \Delta^2 \right]; \\
D_{i} = \mathbb{E}_{\pm\alpha} \mathbb{E}_{\bs{\lambda}}&\left[g(\lambda_i + b_i)  \Delta \right]; \,\,
E_{i} = \mathbb{E}_{\pm\alpha} \mathbb{E}_{\bs{\lambda}} \left[ g'(\lambda_i+b_i) \Delta \right],
\end{align*}
allowing us to write the simple ODEs that characterise the time evolution of the order parameters:
\begin{align} \label{eq:ODEs}
d\,Q_{kl} &= \tilde{\eta} \, (v_k A_{k,l} + v_l A_{l,k}) + \tilde{\eta}^2 v_k v_l B_{kl} \\
d\,M_{k\alpha} &= \tilde{\eta} \, v_k A_{k,K+\alpha} \\
 d\,v_{k} &= \tilde{\eta} \, D_k \\
 d\,b_k &= \tilde{\eta} \, v_k E_k.
\end{align}
As in \citet{refinetti2021classifying}, we rescaled 
the learning rate with the number of neurons, fixing  $\tilde{\eta} = \eta / \sqrt{K} = 2.5$. The obtained order parameters can then be used to track the trajectory of the typical observables and performance metrics for the neural network, such as the generalisation error and the population loss. We provide a sketch of the derivation of the ODE description in Appendix \ref{app:ODE_derivation}.

\textbf{Lottery ticket effect.} In the XGM setting, the minimal architecture that can achieve optimal generalisation is a 2-layer network with $K=4$ neurons \cite{refinetti2021classifying, arous2023high}. The key requirement for an optimal solution is to achieve perfect \emph{coverage}, \ie each of the centroids should have at least one specialised neuron that aligns with it. In \citet{arous2023high}, in the noiseless limit $\sigma\to0$ and the special case of zero bias $\bs b = \bs 0$, the authors propose a combinatorial analysis for characterising the probability of the dynamics falling into the basin of attraction of the perfect coverage solutions, and identify the initial configuration of the neurons as the discriminating factor. In particular, the learning network is shown not to be able to recover from poor initializations where multiple neurons redundantly focus on the same centroid and one or multiple centroids are ``ignored'' by the network---\eg as in Fig.~\ref{fig:fig1}a.\looseness=-1 

As first observed in \citet{refinetti2021classifying}, overparameterising the network by increasing $K$ lowers the probability of the initial configuration being in the basin of attraction of a low coverage solution---as sketched in Fig.~\ref{fig:fig1}b---and leads to a clear improvement in the average performance of the network. In fact, the probability of reaching an optimal solution is shown to increase exponentially with $K$ in the simplified setting of \citet{arous2023high}. We extend the analysis to the case of $\sigma>0$ and non-zero bias, where the network has the additional option of \emph{muting} a neuron by sufficiently increasing the associated bias. In Fig.~\ref{fig:fig1}c, we show the gain in performance for increasing values of $K$ as a function of the noise level $\sigma$. 
Note that the curves, obtained through the ODE system Eq.~\ref{eq:ODEs}, display the averages over initialisations of the population loss reached after $10,000$ learning epochs. At very large values of $\sigma$, the selected cut-off on the epochs is insufficient for the network to converge to a configuration with better-than-chance performance. 

This overparameterisation phenomenology provides a clear example of the lottery ticket hypothesis \cite{frankle2018lottery}. The idea is that one of the advantages of training highly overparameterised neural networks comes from the increased likelihood of randomly sampling a well-initialised sub-network that is sufficient to solve the learning problem at hand. 
In simple words, collecting more lottery tickets will certainly enhance the chance of finding the winning one. 
In the following sections, we investigate the effectiveness of curriculum learning in the XGM problem and whether it can work in synergy with the overparameterisation strategy.

\section{Curriculum learning} \label{sec:curriculum}

\begin{figure*}[h]
    \vskip 0.2in
    \begin{center}
        \begin{subfigure}[b]{0.45\textwidth}
            \centering
            \includegraphics[width=\textwidth]{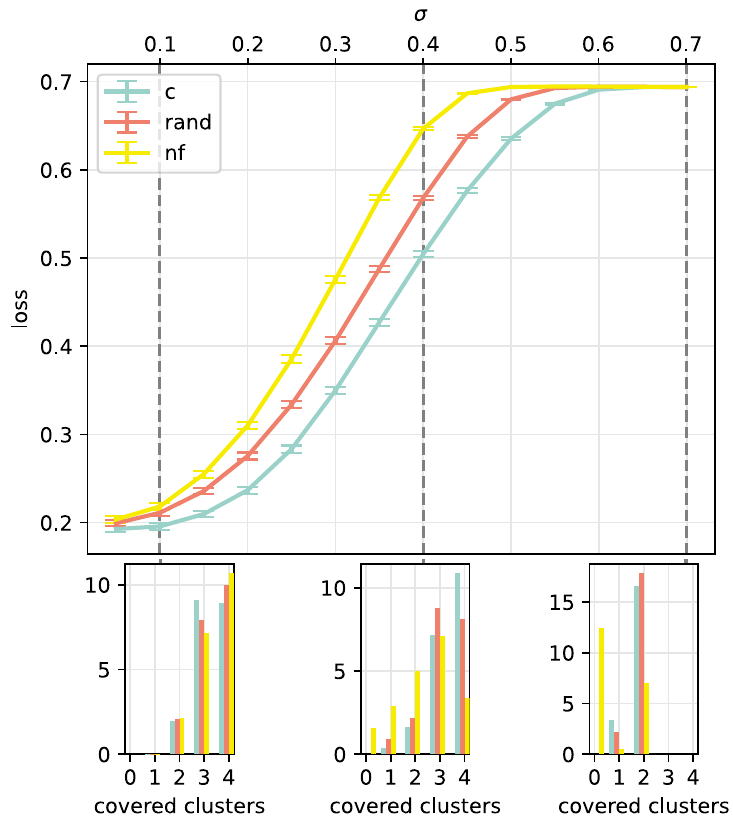}
            \vspace{0.1cm}
            \caption{Asymptotic performance of curricula}
        \end{subfigure}
        \begin{subfigure}[b]{0.54\textwidth}
            \centering
            \includegraphics[width=\textwidth]{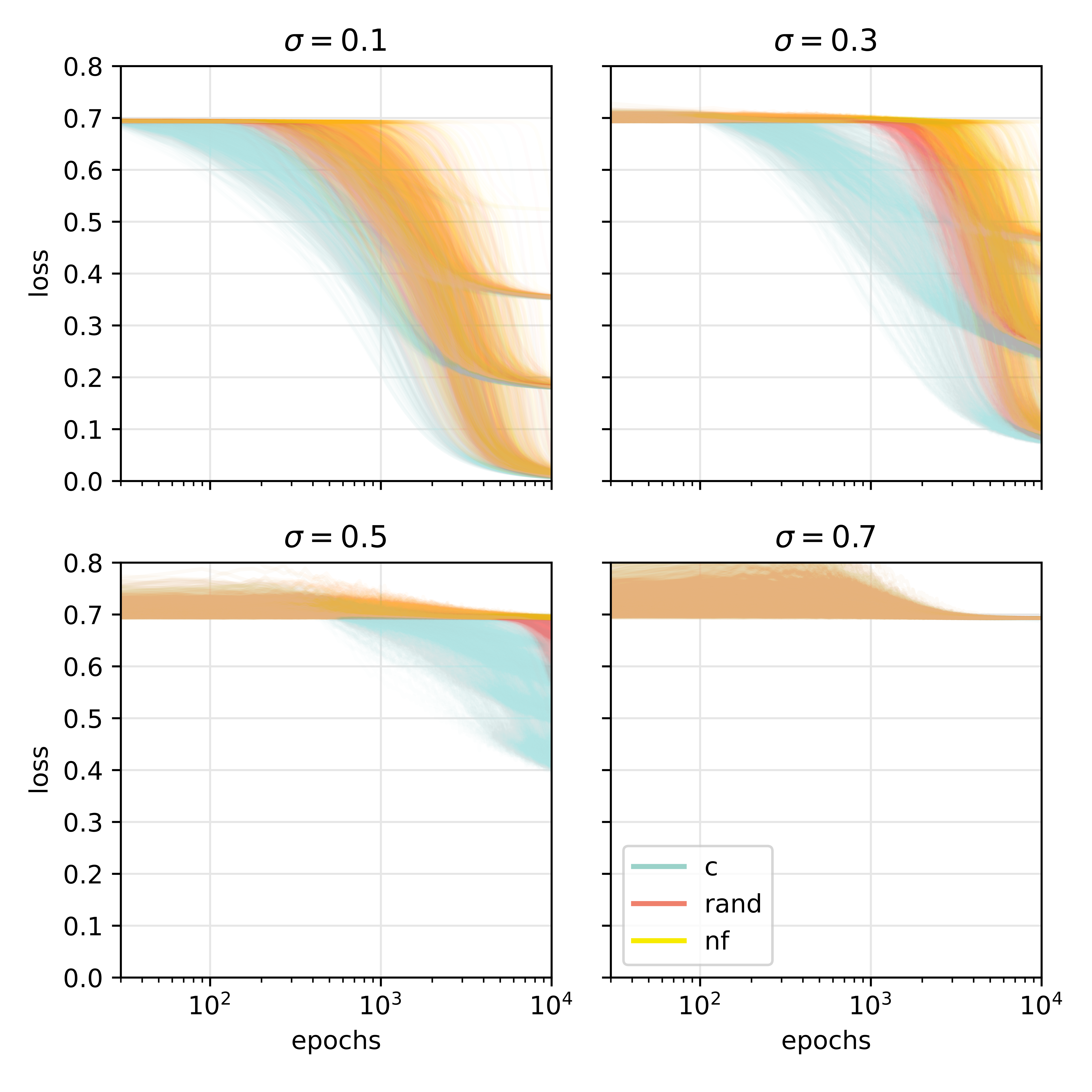}
            \caption{Curricula dynamics}
        \end{subfigure}
    \end{center}
    \caption{
    \textbf{The benefits of a curriculum.}
    \textit{Panel (a)} shows that curriculum learning (`c') can achieve a better test loss in comparison with random ordering (`rand') and no-fading (`nf') paradigms. Consistent with results in cognitive science, neural networks find benefits from curricula in a Goldilocks range of the noise. Histograms show centroid coverage for the three protocols.
    \textit{Panel (b)} highlights another aspect of curricula that comes from the dynamics: curricula can also speed up learning considerably.
    }
    \label{fig:fig2}
\end{figure*}

In curriculum learning, samples are presented to the model in a curated order. This procedure finds motivation in studies from the field of cognitive science \cite{lawrence1952transfer,baker1954discrimination,skinner1965science} showing that animals learn faster---and sometimes learning is unlocked---if samples are presented in increasing order of difficulty. 
However, with few exceptions, deep learning does not seem to benefit as much from curricula \cite{wu2020curricula}. 

The very definition of what identifies a \textit{difficult sample} in standard ML problems is debatable, and in practice, different approaches have been tested with alternating results.
In the synthetic framework of the XGM, we can use the $\mathrm{SNR}=\frac{\lVert\bs\mu\rVert}{\sigma}$ as a natural notion of sample difficulty. In particular, to similar effect, one could either rescale the norm of the centroids $\bs\mu_\alpha$, or the standard deviation $\sigma$, in a fraction of the presented patterns. In the following, we will employ the former approach, often referred to as \emph{fading} in the field of cognitive science \cite{pashler2013does}. To this end, we define a \textit{fading factor} $\varphi$ to rescale the signal component of the inputs $\bs X^{\pm\alpha} \sim \mathcal{N}(\pm \varphi \bs\mu_\alpha, \sigma^2\mathbb{I})$.

We compare three learning protocols, namely \emph{curriculum}, \emph{random order}, and \emph{no-fading}. In the first two cases, we assume that a fraction $\alpha$ of the dataset contains easier samples---\ie $\varphi>1$. 
In the \emph{curriculum} setting, we consider a fading factor that, starting from a maximum value, decreases linearly to $1$ in the first fraction $\alpha$ of the epochs, and is kept fixed thereafter. We set $\varphi_{\max}=3$ in our experiments. In the \emph{random order} protocol, the dataset is presented with identical yet shuffled fading factors. Finally, in the \emph{no-fading} protocol, $\varphi=1$ for all samples---\eg as in Fig.~\ref{fig:fig1}c. Notice that only the \emph{no-fading} protocol is explicitly receiving less information.

\begin{figure*}[h]
    \vskip 0.2in
    \begin{center}
        \begin{subfigure}[b]{0.38\textwidth}
            \includegraphics[width=\textwidth]{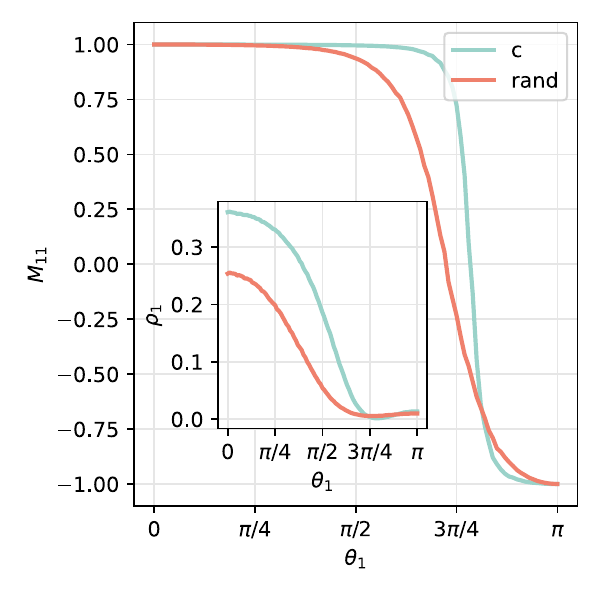}
            \caption{Curriculum benefit}
        \end{subfigure}
        \begin{subfigure}[b]{0.45\textwidth}
            \includegraphics[width=\textwidth]{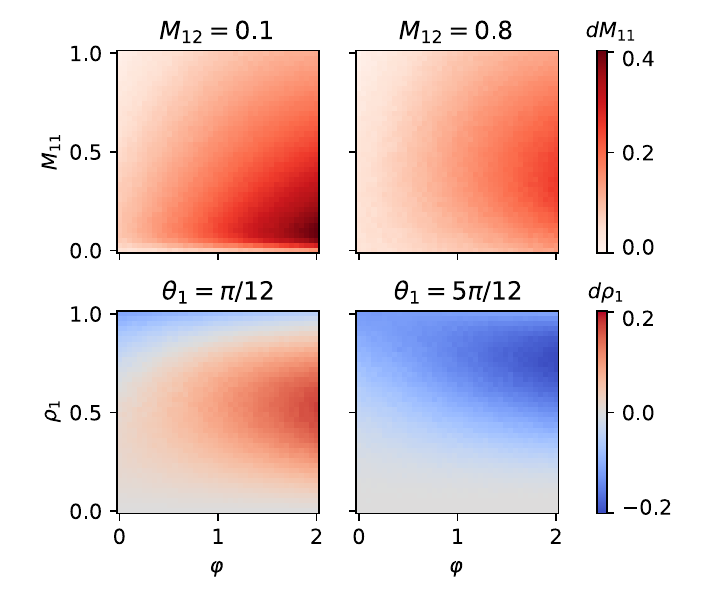}
            \vspace{-0.2cm}
            \caption{Speed boost from curriculum}
        \end{subfigure}
    \end{center}
    \caption{
    \textbf{Scrutinising the curriculum gain.}
    \textit{Panel (a)} shows a controlled experiment where a $K=4$-network is initialised with 3 neurons in the optimal configuration, and the remaining one at an angle $\theta_1$ with the free centroid and with a fraction $\rho_1=0.1$ of norm lying on the 2-dimensional relevant manifold. The experiment is run in the high-noise regime, $\sigma=1.0$. The curriculum (blue) and random (red) lines show that \emph{curriculum} finds the optimal configuration for a significantly larger range of initial angles. Furthermore, the inset shows that the final $\rho_1$ of the free neuron is consistently larger for \emph{curriculum}.
    \textit{Panel (b)} shows the instantaneous benefit of curricula, both for the rate of alignment with the left-out centroid---$d\,M_{11}$ (upper row)---and in the rate of alignment with the relevant manifold---$d\,\rho_1$ (lower row). The heatmaps span the fading factors $\mu$ on the horizontal axis, and the current mass on the relevant manifold $\rho_1$ (alignment with the centroid $m_1$) on the vertical axis of the upper (lower) row. Crucially, at a given $\mu$, the benefit is non-monotonic on the other variable, implying that there is an optimal period during learning for using curricula effectively.
    }
    \label{fig:fig3}
\end{figure*}

\subsection{Effectiveness of curriculum} 

Before analysing the mechanisms behind the effectiveness of curriculum, we here show the twofold nature of the curriculum benefit. For simplicity, we focus on the minimal 2-layer network setting ($K=4$) in the XGM task, with $10,000$ training epochs and an easy fraction $\alpha=0.1$.

In Fig.~\ref{fig:fig2}a and Fig.~\ref{fig:fig2}b, we show that a \emph{curriculum} strategy can provide both an asymptotic and a dynamical advantage compared to \emph{random order} and \emph{no fading}. The upper plot of Fig.~\ref{fig:fig2}a displays the population loss after training, for different levels of standard deviation $\sigma$, while the histograms on the bottom show the cluster coverage achieved by the three strategies at $\sigma\in[0.1,0.4,0.7]$. In Fig.~\ref{fig:fig2}b, we show the learning trajectories (time evolution of the loss) of the different protocols at different noise levels, $\sigma\in[0.1,0.3,0.5,0.7]$. Overall we identify three regimes of effectiveness (dashed vertical lines in Fig.~\ref{fig:fig2}a), depending on the noise level: high $\mathrm{SNR}$, the Goldilocks region at intermediate $\mathrm{SNR}$, and low $\mathrm{SNR}$.

\textbf{High SNR.} As can be seen in Fig.~\ref{fig:fig2}a, at small $\sigma$ all the protocols can identify the relevant manifold spanned by the centroids and progressively align the available neurons to it. However, by providing easier samples at the beginning of the training, \emph{curriculum} lowers the entropic barrier associated with the search of the relevant manifold, allowing for a faster specialisation of the neurons---see also Fig.~\ref{fig:fig2}b---and inducing a quicker growth of the norm of the neurons, which yields a lower loss value on average. Interestingly, the left histogram shows that despite the lower loss, the associated level of coverage of \emph{curriculum} is found to be slightly worse than \emph{random order} and \emph{no-fading}. While a small amount of noise can help escape bad initialisations, the initial boost in the $\mathrm{SNR}$ of the fading strategy commits the network to the closest bad minimum and prevents positive fluctuations, proving detrimental in this regime. 

\textbf{Goldilocks SNR.} At intermediate values of $\sigma$ we find the Goldilocks regime of \emph{curriculum}, where the advantage is maximal. In Fig.~\ref{fig:fig2}a, \emph{curriculum} achieves lower loss and better coverage (central histogram) on average compared to the other strategies. This signals that the fading procedure, by temporarily increasing the $\mathrm{SNR}$, can effectively enlarge the basin of attraction of neuron configurations with high coverage. Fig.~\ref{fig:fig2}b also shows that the learning speed is maximal in this regime, and that larger fractions of trajectories land on lower loss values.

\textbf{Low SNR.} Finally, at large $\sigma$ the complexity of the task is too high for curricula to be effective. The time scale for identifying the relevant manifold spanned by the cluster centroids is large compared to the training epochs, and the model predictions are dominated by the irrelevant---non-informative---dimensions. Thus, the observed performance is nearly chance-level. As shown in the right histogram, a poor alignment with the relevant manifold can push the network towards configurations where all the neurons are muted and none of the clusters is covered.

This preliminary analysis shows the emergence and interplay between two key ingredients: 
\begin{itemize}
\item the \textit{relevant manifold discovery}, i.e. how much the network aligns its neurons to the sub-manifold spanned by the centroids, decreasing the impact of the noise dimensions on the learning dynamics. This sub-task does not explicitly require the labels and is akin to feature discovery in unsupervised settings. 
\item the \textit{labelling rule identification}, i.e. how much the learned features allow the network to represent the correct input-output association for the presented data. In the XGM model, this goal is in direct correspondence with cluster coverage. This sub-task is intrinsically driven by supervision.
\end{itemize}
In the next section, we disentangle these sub-tasks in controlled experiments and show the impact of the presentation order of the $\mathrm{SNR}$-boosted patterns.

\subsection{Anatomy of curriculum} 
We now aim to quantify the impact of \emph{curriculum} on the \textit{relevant manifold discovery} and \textit{labelling rule identification} sub-tasks.
To this end, we consider a controlled setup where most of the network parameters are ideally initialised: $3$ of the $4$ neurons start aligned with distinct centroids, $\bs v$ is chosen with appropriate alternating signs, and $\bs b = \bs 0$. 

The idea is to explore in isolation the impact of the initialisation of the weights of the remaining neuron---neuron $1$ for simplicity. We parameterise it by two numbers: $\theta_1 = \arctan \frac{M_{12}}{M_{11}}$, the angle between the projection of the weights $\bs W_1$ on the relevant manifold and the left-out centroid; and $\rho_1 = \frac{M_{11}^2 + M_{12}^2}{Q_{11}}$, the fraction of norm of the neuron lying on the relevant manifold. 
Ideally, the manifold discovery sub-task would require the neuron to obtain $\rho_1 \to 1$, while the optimal cluster coverage would be obtained when $\theta_1\to0$. When $\rho_1$ is small, the response of the neuron is more affected by noise and its fluctuations are enhanced. For $\rho_1\to1$ the fluctuations weaken, since the signal component becomes more prominent, and the neuron is attracted to the closest centroid compatible with the initialisation of $\bs v$. Note that, despite the semi-aligned initialisation, during training all neurons are left free to evolve and can suffer slight dis-alignments from the relevant manifold---especially in the low $\mathrm{SNR}$ regime.

\textbf{Labelling rule identification.} In Fig.~\ref{fig:fig3}a, we show the final alignment with the left-out centroid, $M_{11}$, after $10,000$ training epochs and starting from $\rho_1=0.1$ and $\theta_1\in[0,\pi]$. The \emph{curriculum} protocol induces a sharper behaviour: a critical value of $\theta_1$ separates a phase where the neuron consistently aligns to the centroid, from a phase where it consistently anti-aligns. 
In contrast, \emph{random order} shows a less consistent behaviour, with a larger fraction of initialisations neither aligning nor anti-aligning with the left-out centroid. Overall, \emph{curriculum} is found to infer the correct labelling rule for a wider range of $\theta_1$s. Interestingly, at large $\theta_1$, \emph{random order} displays a small advantage, connecting back to the observations about the left histogram---$\sigma=0.1$---in Fig.~\ref{fig:fig2}a.
In the inset of Fig.~\ref{fig:fig3}a, we see that after training \emph{curriculum} achieves a larger $\rho_1$ than \emph{random order}, proving more efficient in the discovery of the relevant manifold especially when $\theta_1$ is small.  

By inspecting the ODE updates Eq.~\ref{eq:ODEs}, in Fig.~\ref{fig:fig3}b we can analyse the rate of increase of $M_{11}$ (top) and $\rho_1$ (bottom), as a function of their current values and of the fading factor $\varphi$. 
In the top plots, we study the gradient of the alignment with the left-out centroid, $M_{11}$, at $\rho_1=0.8$ in two different scenarios. 
On the left, the alignment with the
orthogonal centroid is small---$M_{12}=0.1$---and the signal is stronger, while on the right the neuron is more aligned with the orthogonal centroid---$M_{12}=0.8$---increasing the cross-talk between the two directions. The impact of $\varphi$ on the growth rate of $M_{11}$ is found to be maximal when $M_{11}$ is about the same order as $M_{12}$. In a random initialisation scenario, having a stronger signal in the early training stages---when the alignment with both neurons is still small---is thus more beneficial compared to getting the boost in later stages of training. 

\textbf{Relevant manifold discovery.} In the bottom plots of Fig.~\ref{fig:fig3}b, we study the gradient for $\rho_1$ in two scenarios: when the neuron is already well aligned with the left-out centroid ($\theta_1=\frac{1}{12}\pi$, left) and when the alignment is stronger with the orthogonal centroid ($\theta_1=\frac{5}{12}\pi$, right).
At low values of $\varphi$, in both scenarios the gradient is weak. 
However, if $\varphi$ is large we observe a bifurcation: if the neuron is already oriented correctly, it is pulled toward the relevant manifold; if the orientation is incorrect, the neuron is repelled from the relevant manifold. 
This behaviour is reminiscent of a phenomenon exploited in the learning rate warm-up strategy and implicitly in the catapult mechanism \cite{lewkowycz2020large}: temporarily disaligning with the relevant manifold can be useful because the fluctuations induced by the irrelevant directions can be used to escape a sub-optimal basin. See Appendix \ref{app:catapult_effect} for additional details.

Notice that identifying explicit parameters that track the progress in the two sub-tasks is only feasible in controlled settings like the XGM. A similar analysis with real data would require some approximations to estimate analogous observables. For instance, reduction techniques like PCA or CKA~\cite{kornblith2019similarity} could be used to estimate the manifold of relevant features, and heuristic methods like fitting a teacher-network to the data~\cite{loureiro2021learning} could allow a more controlled characterisation of the rule. 

\begin{figure*}[h]
    \begin{center}
        \begin{subfigure}[b]{0.6\textwidth}
            \centering
            \includegraphics[width=\textwidth]{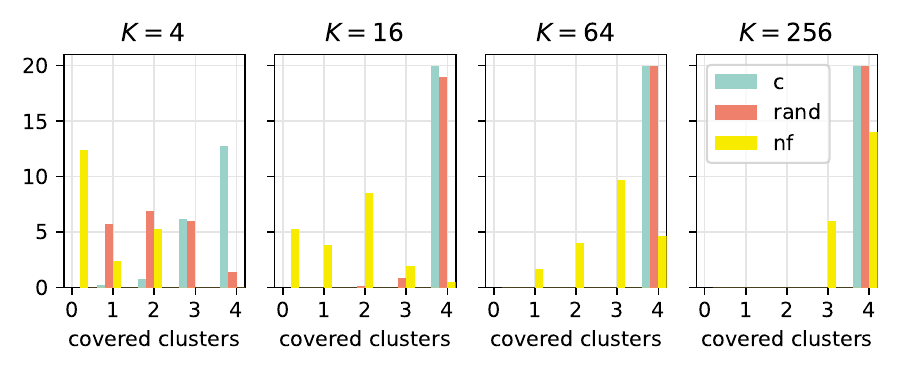}
            \caption{Curriculum in the overparameterised regime}
        \end{subfigure}
    \end{center}
    \begin{center}
        \begin{subfigure}[b]{0.9\textwidth}
            \centering
            \includegraphics[width=\textwidth]{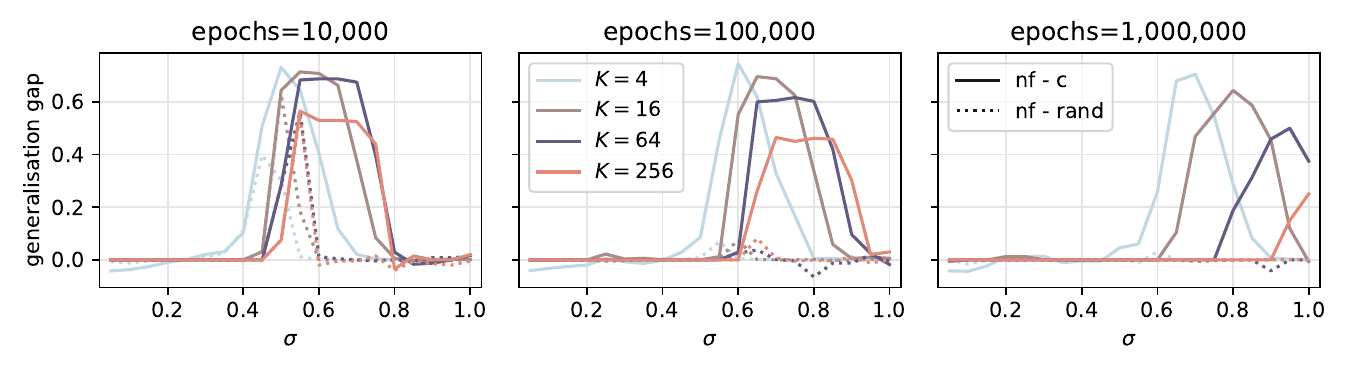}
            \caption{Curriculum performance difference}
        \end{subfigure}
    \end{center}
    \caption{
    \textbf{Interaction between curricula and overparameterisation.}
    \textit{Panel (a)} shows histograms of the cluster coverage achieved after 10,000 epochs, for a standard deviation $\sigma=0.4$ and 1,000 easy samples ($\alpha=0.1$). The colours (blue, red, and yellow) denote the strategies used (\emph{curriculum}, \emph{random order}, and \emph{no-fading}, respectively) during training. The different plots in the panel show the effect of overparameterisation, controlled by the parameter $K$. While a small degree of overparameterisation benefits curricula more than other strategies, a very large overparameterisation makes all strategies equally effective.
    \textit{Panel (b)} generalises the picture to a broad range of $\sigma$s and larger training time. In all these cases, we always keep the number of `easy' samples fixed to 1,000---\eg in the rightmost plot only 0.1\% of the samples are easy. The different lines represent the gap in noiseless generalisation error between \emph{curriculum} and \emph{no-fading} (solid lines), and \emph{random order} and \emph{no-fading} (dotted lines), for different network parameterisations $K$.  
    }
    \label{fig:fig4}
\end{figure*}

\subsection{Ablations and variations}

Our analysis, so far, focused on a specific setting, trying to remove as many elements of complexity as possible, in order to identify the simplest model avoiding confounding effects.
However, practical applications use a number of tricks to improve performance. We discuss some of these changes, showing the robustness of our result, in Appendix~\ref{app:ablations}.

\section{Interplay of curriculum learning and over-parameterization}

We have so far separately covered the benefits of overaparameterisation and curriculum learning. We now turn to characterising their interplay. In particular, we address the question: \textit{How effective can curriculum be when the model approaches the overparameterised regime?}

In Fig.~\ref{fig:fig4}a, we go back to the random initialisation setup, and plot the histograms for the cluster coverage achieved by networks with increasing numbers of neurons and trained according to the three curriculum protocols. 
The results show that both employing \emph{curriculum} and overparameterising the network can increase the probability of ending up in a good solution with high centroid coverage, inducing a lower loss at the end of training.
However, the asymptotic gain of curriculum is minimal when the network is already overparametrised, since full centroid coverage is already achieved without boosting the $\mathrm{SNR}$, and the margin for a curriculum benefit is no longer present. 

\looseness=-1
In Fig.~\ref{fig:fig4}b, we extend the picture by exploring a range of values for $\sigma$ and by increasing the number of training epochs, keeping the total number of easy samples fixed to $1,000$. Each of the plots displays, for different network sizes $K$, the gap between the generalisation error evaluated at $\sigma=0$ of \emph{curriculum} and \emph{no-fading} (full lines), and the gap between \emph{random order} and \emph{no-fading} (dotted lines). First, notice that with larger $K$ the maxima of the curves shift to the right, indicating that higher levels of noise can be tolerated in overparameterised networks, and that the \emph{curriculum} Goldilocks regime shifts to higher $\sigma$.
At fixed $\sigma$, the asymptotic gain of curriculum tends to decrease with larger $K$, as observed in Fig.~\ref{fig:fig4}a.

\begin{figure*}[h!]
    \vskip 0.2in
    \begin{center}
        \begin{subfigure}[b]{0.4\textwidth}
            \centering
            \includegraphics[width=\textwidth]{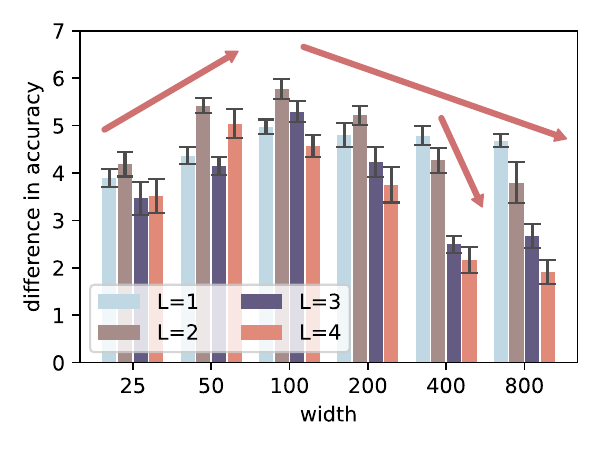}
            \caption{Curricula in MLP}
        \end{subfigure}
        \begin{subfigure}[b]{0.3\textwidth}
            \centering
            \includegraphics[width=\textwidth]{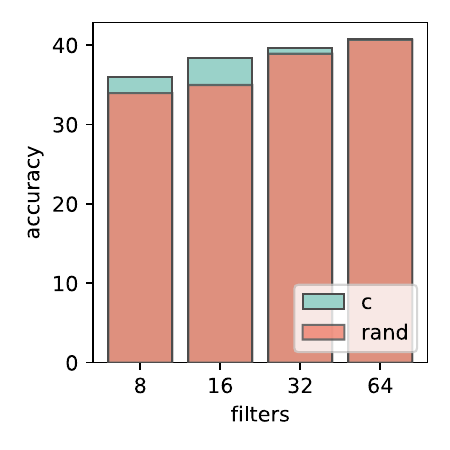}
            \caption{Curricula in CNN}
        \end{subfigure}
        \hspace{0.4cm}
        \begin{subfigure}[b]{0.18\textwidth}
            \centering
            \includegraphics[width=\textwidth]{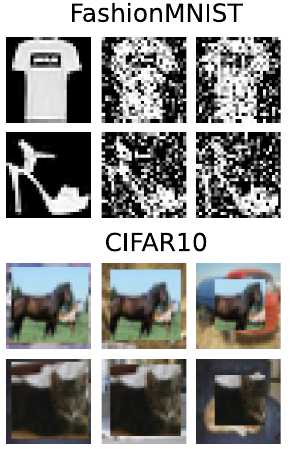}
            \vspace{0.2cm}
            \caption{Dataset samples}
        \end{subfigure}
    \end{center}
    \caption{
    \textbf{Interplay between curriculum and overparameterisation in real data.}
    \textit{Panel (a)} shows the effect of curriculum and overparameterisation for a MLP trained in a corrupted fashionMNIST dataset. 
    We can observe two types of overparameterisation, layer-wise (x-axis) and in depth (bars), and their impact on the gap between \emph{curriculum} and \emph{random order} (y-axis). The key observation is that, above the Goldilocks range, both kinds of overparameterisation reduce the benefit of curricula.
    \textit{Panel (b)} extends the analysis on CNNs trained in a corrupted CIFAR10 dataset. As we overparameterise by increasing the number of filters (x-axis), the accuracies (y-axis) of the two strategies get closer and their gap decreases.
    In \textit{Panel (c)} we show some of the samples used in training and testing. For the MLP, we add `difficulty' to fashionMNIST by adding white noise to the images. Since CNNs are robust against this kind of perturbation, we increase the `difficulty' of CIFAR10 samples by adding a distracting frame around the image. The different columns represent the easy samples, the test samples and the hard samples, respectively.
    }
    \label{fig:fig5}
\end{figure*}

\looseness=-1
With a larger number of training epochs---Fig.~\ref{fig:fig4}b, mid and right plots---the network has sufficient time to converge closer to the minimum within the reached basin of attraction. Therefore, the dynamical speed-up effect of \emph{curriculum} becomes negligible. In these settings, the performance gaps indicate a higher frequency of convergence to solutions with good centroid coverage. We also observe that the maximum gap decreases when $K$ is larger, evidencing again that in the overparameterised regime curriculum strategies have less to gain.

Looking at the difference between \emph{random order} and \emph{no-fading} (dotted curves) we see that it vanishes for longer training times: \emph{random order} struggles to extract additional information from easy samples if they are presented with very low frequency. On the contrary, \emph{curriculum} can still benefit from them. This result can be identified with the \textit{``eureka effect''} described in cognitive science \cite{ahissar1997task,ahissar2004reverse}, where better asymptotic performance can be unlocked if easier examples are consistently presented in the initial stages of learning.

\looseness=-1
Finally, notice that curriculum can cause worse performance in the regime of very small noise and a large number of epochs. This is compatible with the previous observations on Fig.~\ref{fig:fig2}a and Fig.~\ref{fig:fig3}a: since some noise at the beginning of learning can help escape sub-optimal basins of attraction, our \emph{curriculum} protocol can be detrimental since it reduces the fluctuations.

\section*{Numerical results on ML benchmarks}

In this section, we seek a similar phenomenology on real data. As discussed, defining a proper notion of difficulty among the patterns of standard datasets---\eg FMNIST or CIFAR10---is an open problem, and is out of the scope of the present paper. Here, we simply modify the original dataset to complexify the task in a tunable way. Given the limited size of the datasets, we revert to the standard batch-learning setting with cross-entropy loss. Curriculum learning can be introduced by splitting the learning process into stages, where the model can focus on different slices of the dataset.

In the first experiment on FMNIST \cite{xiao2017fashion}, we simply add i.i.d.~Gaussian noise to the input pixels and clip to the original range $[0,1]$. The difficulty level is thus controlled via the standard deviation of the additive noise---see top plot in Fig.~\ref{fig:fig5}c. We consider a sub-sampled dataset with $5,000$ unperturbed `easy' samples and $5,000$ perturbed `hard' samples with $\sigma=0.5$. For \emph{curriculum}, in the first training stage we present to the network only the easy samples, drop the learning rate by a factor $1/3$ and finally train only on the hard samples in the second and final stage. For \emph{random order} we simply train on the full dataset in shuffled order for the same total epochs. We test on hard patterns with noise $\sigma=0.5$. We train with Adam \cite{kingma2014adam} a multi-layer perceptron with $L$ layers, each with the same width.  In Fig.~\ref{fig:fig5}a, we find that, in agreement with the XGM phenomenology, the gain of \emph{curriculum} first increases as the width is increased to $100$---the Goldilocks regime---and then decreases when the degree of overparameterisation is further increased. Also increasing the number of layers has a similar effect---\ie the gap shrinks, especially when the width is sufficiently large.

In the second experiment on CIFAR10 \cite{krizhevsky2009learning}, we introduce a difficulty level by swapping centres and frames of different images in the dataset---maintaining the label associated with the center. The difficulty level is controlled via the width of the distracting frame---see bottom plot in Fig.~\ref{fig:fig5}c. This type of perturbation is designed to be more impactful on convolutional networks. We consider a training set with $1,000$ easy samples with distracting frame width $w=1\,\mathrm{pixel}$  and $9,000$ hard samples with $w=6\,\mathrm{pixels}$, and then test on samples with $w=2\,\mathrm{pixels}$. In the \emph{curriculum} protocol, we first train on the easy samples alone, and then train on the whole dataset. We train with Adam a custom convolutional network---additional details in Appendix \ref{app:real_data}---where the number of filters in each layer is multiplied by an increasing factor, reported in Fig.~\ref{fig:fig5}c. The results are again consistent with our theoretical analysis, supporting the idea that overparameterisation limits the effectiveness of curricula.
\section*{Discussion}

We provided a detailed theoretical characterisation of the effectiveness of curriculum learning in training a 2-layer network on the XOR-like Gaussian Mixture problem. After establishing a connection with recent studies, proving an advantage of overparameterisation in the same setting, we highlighted the limitations for the two strategies to work in synergy, since in the highly overparameterised regime the margin of improvement for curriculum eventually vanishes. A similar phenomenology was then traced in experiments with real data and more complex neural network architectures. 
More generally, overparameterisation was recently shown to smooth and convexify the loss landscape in different settings, \eg in 2-layer neural networks~\cite{chizat2018global,mei2018mean,rotskoff2018trainability} and in phase retrieval \cite{chen2019gradient,sarao2020complex,sarao2020optimization}. At the same time, in a convex setting curriculum learning was shown to be unable to provide a sizeable benefit on asymptotic performance \cite{saglietti2022analytical}.  Combining these observations, we conjecture that curriculum ineffectiveness due to overparameterisation should generalise beyond the particular cases analysed in this paper.

Some deep learning applications, especially in NLP and RL problems \cite{soviany2022curriculum}, show that curricula can be effective in practice for models with very high parameter counts. A possible explanation of this observation is that, despite their size, the trained models might still not be overparameterised enough---compared to the task difficulty---to make curricula ineffective. This observation requires further investigation and is left for future research. 

Recent work in data imbalance~\cite{ye2021procrustean} and fairness~\cite{ganesh2023impact} suggested the use of curricula as mitigation strategies in stochastic gradient descent optimisation. Given the availability of high-dimensional models of fairness~\cite{mannelli2022unfair}, this represents an interesting future direction for our theory. 

Finally, this research also suggests that, when there is external pressure over cost-effectiveness of computation---\eg in biological settings---learning systems that are not too overparametrised can still perform well when curriculum strategies are available. This seems consistent with the crucial role played by curricula in animal learning.

\section*{Impact statement}

The research presented in this paper is fundamentally theoretical, and as of now, we do not foresee substantial direct societal impact stemming from its findings.

\section*{Acknowledgement}
We thank Chris Moore for thought-provoking questions. SSM acknowledges Bocconi University for the hospitality during the final stage of this project.
This work was supported by the Alan Turing Institute (UK-IT Trustworthy AI grant) to SSM, by a Sir Henry Dale Fellowship from the Wellcome Trust and Royal Society (216386/Z/19/Z) to AS, and the Sainsbury Wellcome Centre Core Grant from Wellcome (219627/Z/19/Z) and the Gatsby Charitable Foundation (GAT3755). 


\bibliography{bibliography}
\bibliographystyle{icml2024}

\newpage
\appendix
\onecolumn
\section{Sketch of the derivation of the ODE description.}\label{app:ODE_derivation}

We here provide some additional details on the derivation of the ODE system Eq.\ref{eq:ODEs} for our learning problem --\ie online learning in the XGM problem for a 2-layer network with $K$ neurons. To recover the low-dimensional deterministic description we are going to consider the asymptotic limit where the input size $d\to\infty$. Note that the full derivation was first derived in \cite{refinetti2021classifying}, and then rigorously proven in \cite{arous2023high}.

The network is trained using one-pass stochastic gradient descent on a given loss function. Note that, while in this work we focus on the case of binary cross-entropy loss $L(y,\hat{y}) = -y \hat{y} + \log\left( 1 + \exp(\hat{y})\right)$ --as in \cite{arous2023high}, a different loss could be considered with small adjustments --\eg a MSE loss $MSE(y,\hat y) = (\hat y - y)^2/2 = \Delta^2/2$ was analysed in \cite{refinetti2021classifying}. 

Given an input-output pair $(X,y)$, we define the receptive field of the neurons $\lambda_k = \bs W_k \cdot \bs X$, and the ``error'' $\Delta = y-\sigma\left(\frac{1}{\sqrt{K}} \sum_k v_k g(\lambda_k + b_k)\right)$, to write the discrete time SGD updates:
\begin{align}
     &\bs W_k^{t+1} = \bs W_k^t - \frac{\eta}{\sqrt{K}} \Delta \, v_k g'\left( \lambda_k + b_k \right) \bs X
     \\
     &b_k^{t+1} = b_k^t - \frac{\eta}{\sqrt{K}} \Delta \,v_k g'\left( \lambda_k + b_k \right)
     \\
     &v_k^{t+1} = v_k^t - \frac{\eta}{\sqrt{K}} \Delta \, g\left( \lambda_k^\mu + b_k \right),
\end{align}
where $\eta=\mathcal{O}(1)$ represents the learning rate.

As described in the main text, we want to track the evolution of the order parameters:
\begin{equation}
Q = \bs W \bs W^\top, \,\,\,\, M_{\cdot \alpha} = \bs W \bs \mu_\alpha, \,\,\,\, T_{\alpha \beta} = \bs\mu_\alpha \cdot \bs\mu_\beta.
\end{equation}
By substituting the SGD update equations for the first layer weights we get:
\begin{equation*}
     Q_{jk}^{t+1} = Q_{jk}^{t} + \bs W_j^t \cdot d \bs W_k + (j\leftrightarrow k) + d \bs W_j \cdot d\bs W_k
\end{equation*}
\begin{equation}
     = Q_{jk}^{t} + \frac{\eta}{\sqrt{K}} \, v_k \left( \lambda_j g'\left( \lambda_k + b_k \right) \Delta \right) + (j\leftrightarrow k) + \frac{\eta^2 \sigma^2}{K} v_j v_k \left( g'\left( \lambda_k + b_k \right) g'\left( \lambda_j + b_j \right) \Delta^2 \right),
\end{equation}
and in expectation over the receptive fields:
\begin{equation}
     dQ_{jk} = \frac{\eta}{\sqrt{K}} \, v_k \,\mathbb{E}[\lambda_j g'\left( \lambda_k\right) + b_k \Delta] + (j\leftrightarrow k) + \frac{\eta^2 \sigma^2}{K} v_j v_k \,\mathbb{E}[g'\left( \lambda_k + b_k \right) g'\left( \lambda_j + b_j \right) \Delta^2 ].
\end{equation}
Similarly, we find:
\begin{equation*}
     M_{k\alpha}^{t+1} = M_{k\alpha}^{t} + \bs \mu_\alpha \cdot d\bs W_k 
\end{equation*}
\begin{equation}
     = M_{k\alpha}^{t} + \frac{\eta}{\sqrt{K}} v_k \,\mathbb{E} [\left( \bs \mu_\alpha \cdot \bs X \right) g'\left( \lambda_k + b_k \right) \Delta]
\end{equation}
Following the unified notation introduced in the main text, we introduce $2$ additional fields $\lambda_{k+\alpha} = \bs \mu_\alpha \cdot \bs X$, representing the projection onto the centroid directions of the inputs. 

In the high-dimensional limit $d\to\infty$, assuming i.i.d. inputs from the XOR-like Gaussian mixture, the vector field $\bs \lambda$ becomes Gaussian distributed. To identify the associated mean and covariance, we notice that one can decompose:
\begin{equation}
    \bs W_k = M_{k1} \bs \mu_1 + M_{k2} \bs \mu_2 + \bs W_k^\perp,
\end{equation}
isolating relevant and irrelevant components of the weights of the neuron. Recalling that $\bs X^{\pm\alpha} = \pm \bs\mu_\alpha + \bs z$, with i.i.d. noise components $z_i \sim \mathcal{N}(0,\sigma^2)$, one can easily find that:
\begin{equation}
    m = [ \pm M_{\cdot \alpha}, T_{\cdot\alpha} ]; \,\,\,\,\,\,C = \sigma^2 \begin{bmatrix}Q & M \\ M^\top & T \end{bmatrix};
\end{equation}
where $T$ is the overlap matrix between the centroids, \ie a $2\times 2$ identity matrix in the XGM setting. After the definition of the expectations $A$, $B$, $D$ and $E$, we finally obtain the system Eq.\ref{eq:ODEs}. Note that, in the main text we introduce a rescaled learning rate $\tilde{\eta} = \frac{\eta}{\sqrt{K}}$, which is kept fixed to avoid a trivial slow-down of the learning dynamics in the large $K$ regime. 

\section{Exploiting noise to achieve better alignment.} \label{app:catapult_effect}

\begin{figure}[ht!]
    \begin{center}
        \includegraphics[width=0.85\textwidth]{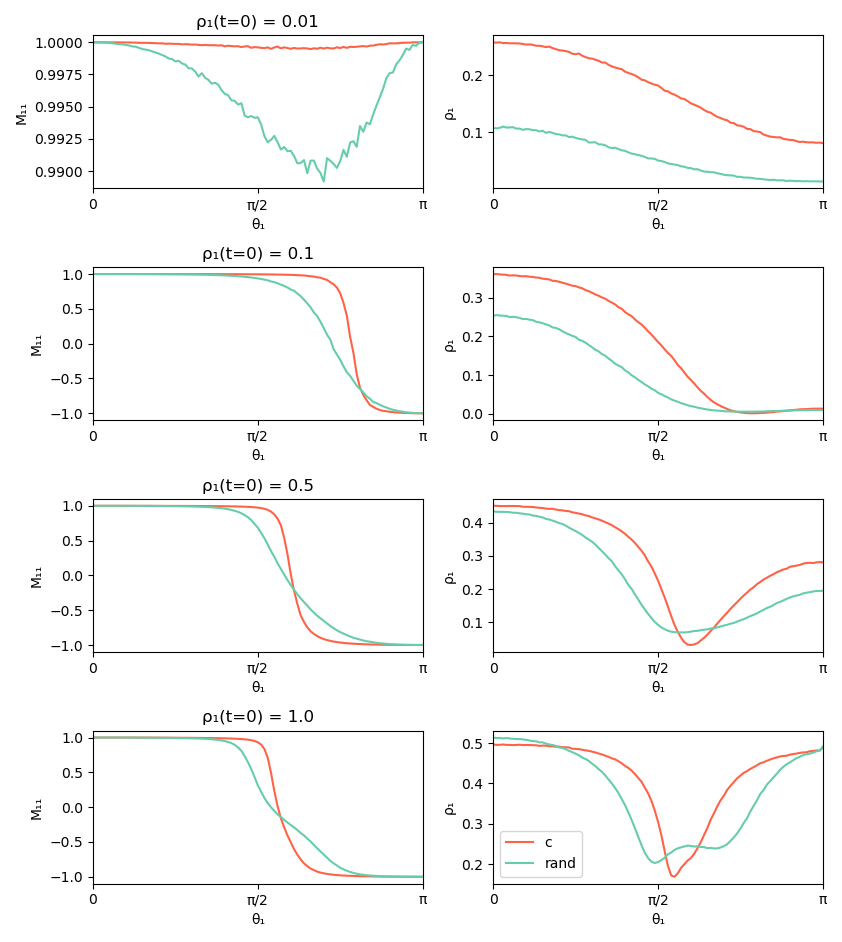}
    \end{center}
    \caption{\label{fig:appB}
    \textbf{More on the two learning tasks in the controlled setup.}
    In each row, we display analogous plots as those in Fig.\ref{fig:fig3}a and its inset, but with an initialising with different initial values of $\rho_1$. In particular, we plot the final alignment of the free neuron with the left-out centroid (left) and the final fraction of relevant norm $\rho_1$, as a function of the initial angle $\theta_1$ between the neuron and the centroid.
    }
\end{figure}

We here present some additional results obtained in the controlled setup introduced in Sec.\ref{sec:curriculum}. 

In particular, in Fig.\ref{fig:appB} we investigate the role played by the initial relevant fraction of the norm $\rho_1$. We observe that, somewhat counterintuitively, initialising neuron $1$ closer to the relevant manifold makes the adversarial initialisations with $\theta_1 \simeq \pi$ more difficult to escape. Instead, when the neuron is initialised with a very small relevant component, the initial $\theta_1$ is almost irrelevant, and the left-out centroid is always covered at the end of training. This is evidence of a mechanism where the noise can be exploited to escape the basin of attraction of sub-optimal minima.

Another observation is that, due to the high level of noise $\sigma=1$, even when the neuron is initialised on the relevant manifold, the learning dynamics causes the irrelevant component of the norm to grow, causing a partial dis-alignment from the relevant manifold. The observed dis-alignment is largest around the $\theta_1$ threshold where the neuron transitions from focusing on the left-out cluster to focusing on the opposite cluster.  

\section{Ablations and variations.}\label{app:ablations}

\begin{figure*}[h]
    \vskip 0.2in
    \begin{center}
        \begin{subfigure}[b]{0.33\textwidth}
            \includegraphics[width=\textwidth]{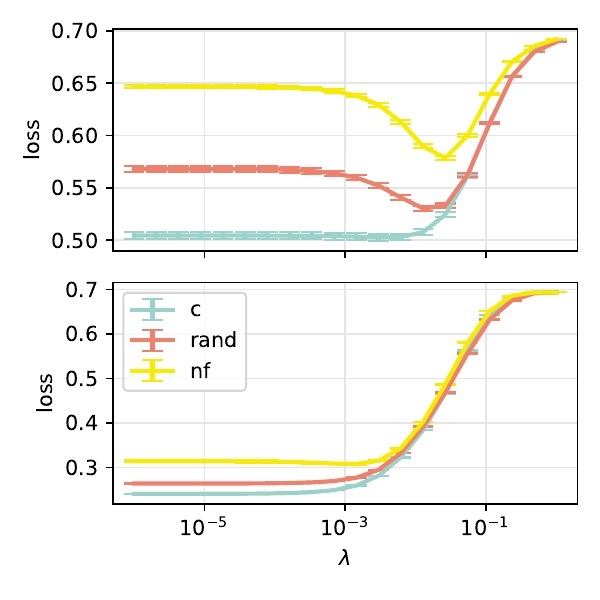}
            \caption{Regularisation}
        \end{subfigure}
        \begin{subfigure}[b]{0.50\textwidth}
            \includegraphics[width=\textwidth]{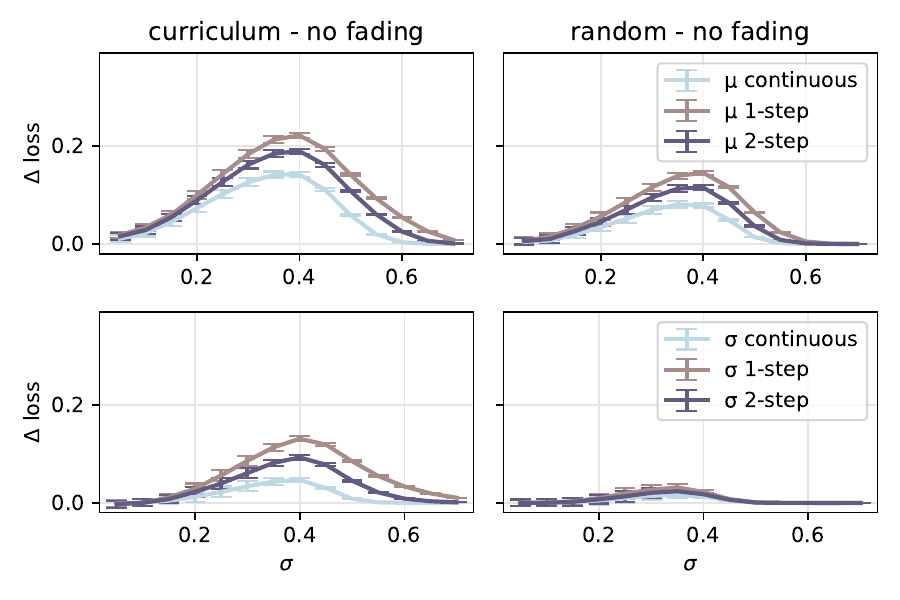}
            \vspace{-0.3cm}
            \caption{Alternative definitions of hardness}
        \end{subfigure}
    \end{center}
    \caption{
    \textbf{Variations of the setting.}
    \textit{Panel (a)} shows the effect of an L2-regulariser with strength $\lambda$ applied to the first layer. Given $\sigma=0.4$, the upper plot reports results for $K=4$ while the lower plot shows the effect on $K=64$. The same analysis with regularisation applied to both layers reported minor benefits.
    \textit{Panel (b)} shows the benefit of curricula against no fading (left plots) and random ordering against no fading (right plots) under alternative notions of difficulty. In the upper plots difficult samples are defined in terms of the norm of the centroids---easy samples have up to twice the norm---while in the lower plots difficulty comes from the cluster variance---with easiest difficulty level given by $\sigma=0.1$. The lines represent continuous curricula, and discrete curricula with 1 or 2 step difficulty increase. 
    }
    \label{fig:fig-rebuttal}
\end{figure*}

Standard optimisation usually includes learning rate scheduling. In general, learning rate annealing advantages curricula. Indeed an effective curriculum protocol can exploit learning rate annealing to focus on relevant samples and prevent forgetting as difficulty increases~\cite{zhou2021curriculum}. 
Another standard protocol is regularisation, which indeed can help identify the relevant manifold by pruning irrelevant directions. In Fig.~\ref{fig:fig-rebuttal}a we show the effect of regularisation in our problems in the just-right parameterised (upper plot) and overparameterised (lower plot) architectures. The plots show an improvement in performance that benefits mostly ``no-fading" and ``random" strategies. This is to be expected as curriculum is already helping in identifying the relevant manifold so the benefit of regularisation is reduced. As shown in the lower plot, the benefit in the overparemeterised architecture is minor, and this is due to the lottery ticket effect, as explained in detail in the next section.

In our investigation we considered curricula based only on one notion of difficulty. In general, defining what characterises a hard samples is not straightforward and it is one of the main design choices behind a curriculum strategy~\cite{soviany2022curriculum}. In Fig.~\ref{fig:fig-rebuttal}b, we investigate several variations of difficulty that still retain analytical tractability. Similarly to~\citet{lawrence1952transfer,baker1954discrimination}, we consider curricula that are discrete---where difficulty assumes 2 or 3 values---and continuous---where there is a continuous increasing in difficulty. Finally, we consider two notions of difficulty based on the centroids' norm $\mu$---similar to the idea of fading used in~\citet{pashler2013does} where relevant features are highlighted---and based on the level of noise in the input $\sigma$---as in~\citet{bengio2009curriculum,saglietti2022analytical}. Overall, despite the quantitative difference, we observe the same behaviour as in the analysis reported so far.

\section{Additional details on the loss in the XOR models.}
\label{app:moreXOR}

\begin{figure}[ht!]
    \begin{center}
        \includegraphics[width=0.4\textwidth]{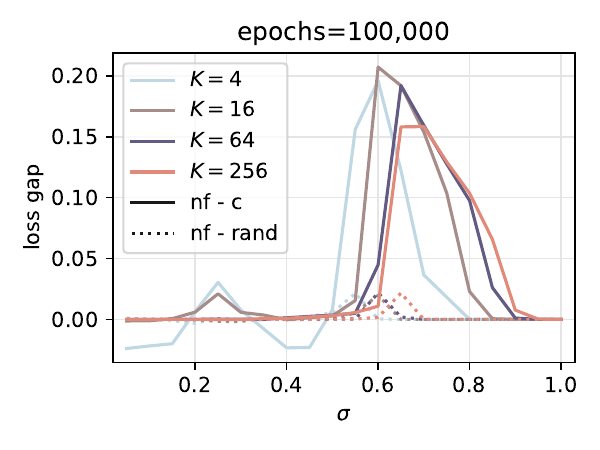}
    \end{center}
    \caption{\label{fig:ap-XOR-loss}
    \textbf{Loss and overparameterisation.}
    The figure shows the loss gap as function of noise $\sigma$, for different level of overparametereistion (in colours). As in Fig.~\ref{fig:fig4}, solid line represent the gain of curriculum versus no-fading while dotted is the gain of random.
    }
\end{figure}

In the main content of the paper we showed that the XOR-like Gaussian Mixture model allows to exemplify a problem that can occur in more complex architectures and tasks. In this section, we want to comment further on some of the peculiarities of this model that however we do not believe share the same degree of universality. 

In particular the actual loss of on the XGM is determined by two main factors: cluster coverage, norm of the layers. In our main discussion we focused mostly on cluster coverage as this is a proxy for how well the network is reconstructing the rule. However, measuring directly the loss can give results harder to interpret. 
In Fig.~\ref{fig:ap-XOR-loss} we show an analogous version of Fig.~\ref{fig:fig4}b measuring the gap in cross-entropy loss rather than generalisation error. While the overall message remains unchanged, \ie overparameterisation makes curricula ineffective, we can notice the presence of an additional bump and a dip for small $K$. 
These are due to a destabilisation of fixed points that occur at different level of SNR for the different strategies.

\begin{figure}[ht!]
    \begin{center}
        \includegraphics[width=0.9\textwidth]{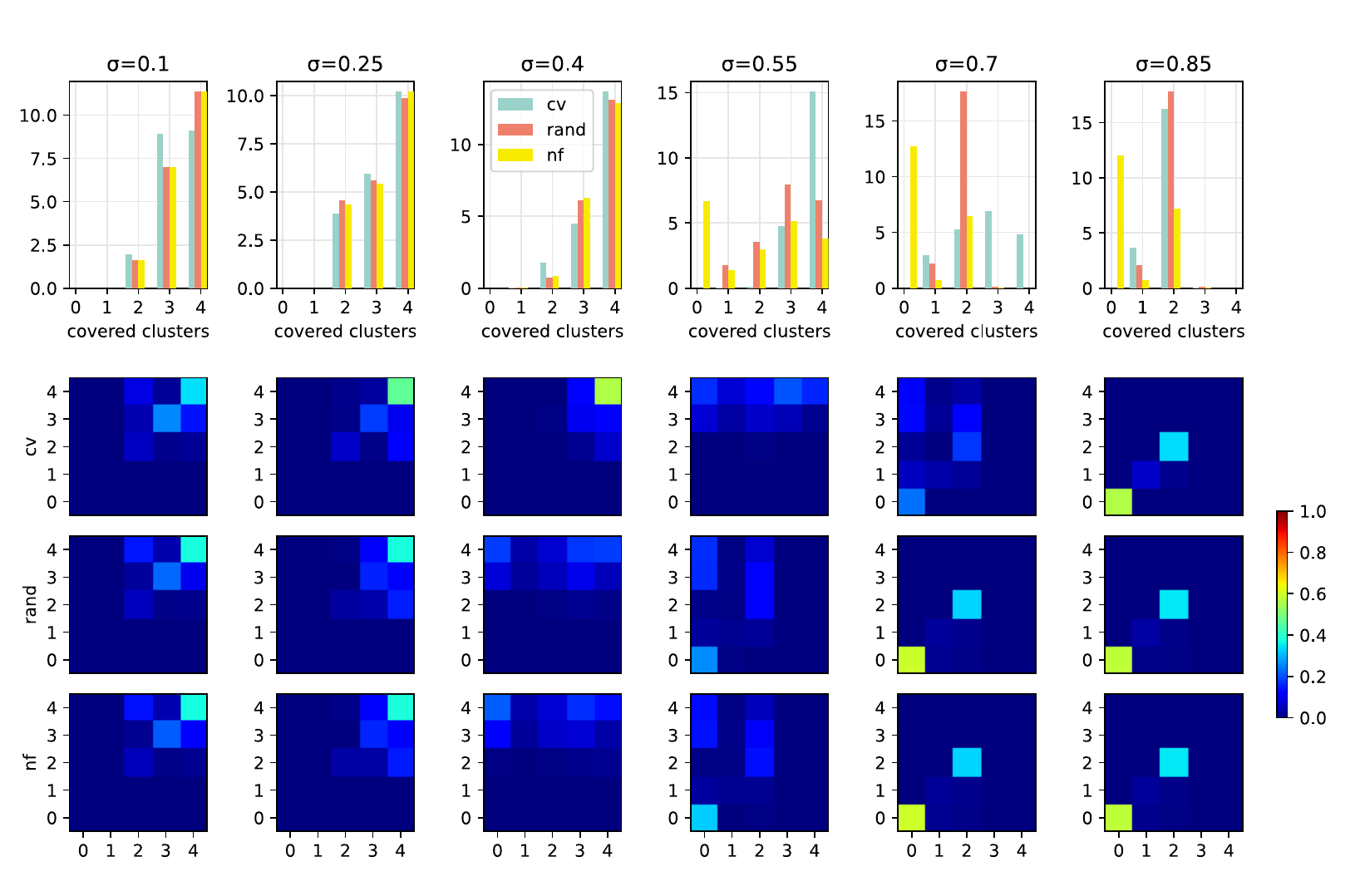}
    \end{center}
    \caption{\label{fig:ap-XOR-transitions}
    \textbf{Fixed point destabilisation.}
    The figure shows the cluster coverage at different values of the noise, $\sigma\in\{0.1,0.25,0.4,0.55,0.7,0.85\}$, indicated by the columns. The first row shows the cluster coverage at given SNR, the following rows instead show the fractions of simulation that move a given level of cluster coverage to another according to their strategy, curriculum, random, no-fading, respectively.
    }
\end{figure}

In order to investigate this further we analysed how the different initialisation get destabilised by increasing the noise. This is shown in Fig.~\ref{fig:ap-XOR-transitions}, where on the first row we show the cluster coverage for the three strategies (respectively curriculum, random, no-fading) at different value of noise. The following three rows show the fraction of initialisations that get destabilised by increasing the noise by $\Delta\sigma=0.15$ and move from a given level of cluster coverage (row of the matrices) to another given level (column of the matrices). Notice, that the pattern between random and no-fading (bottom two rows) is very consistent, while curriculum (2nd row) appears to be delayed. 
This results in a larger stability of fixed points in the curriculum strategy and consequently better performance. 

\begin{figure}[ht!]
    \begin{center}
        \includegraphics[width=0.88\textwidth]{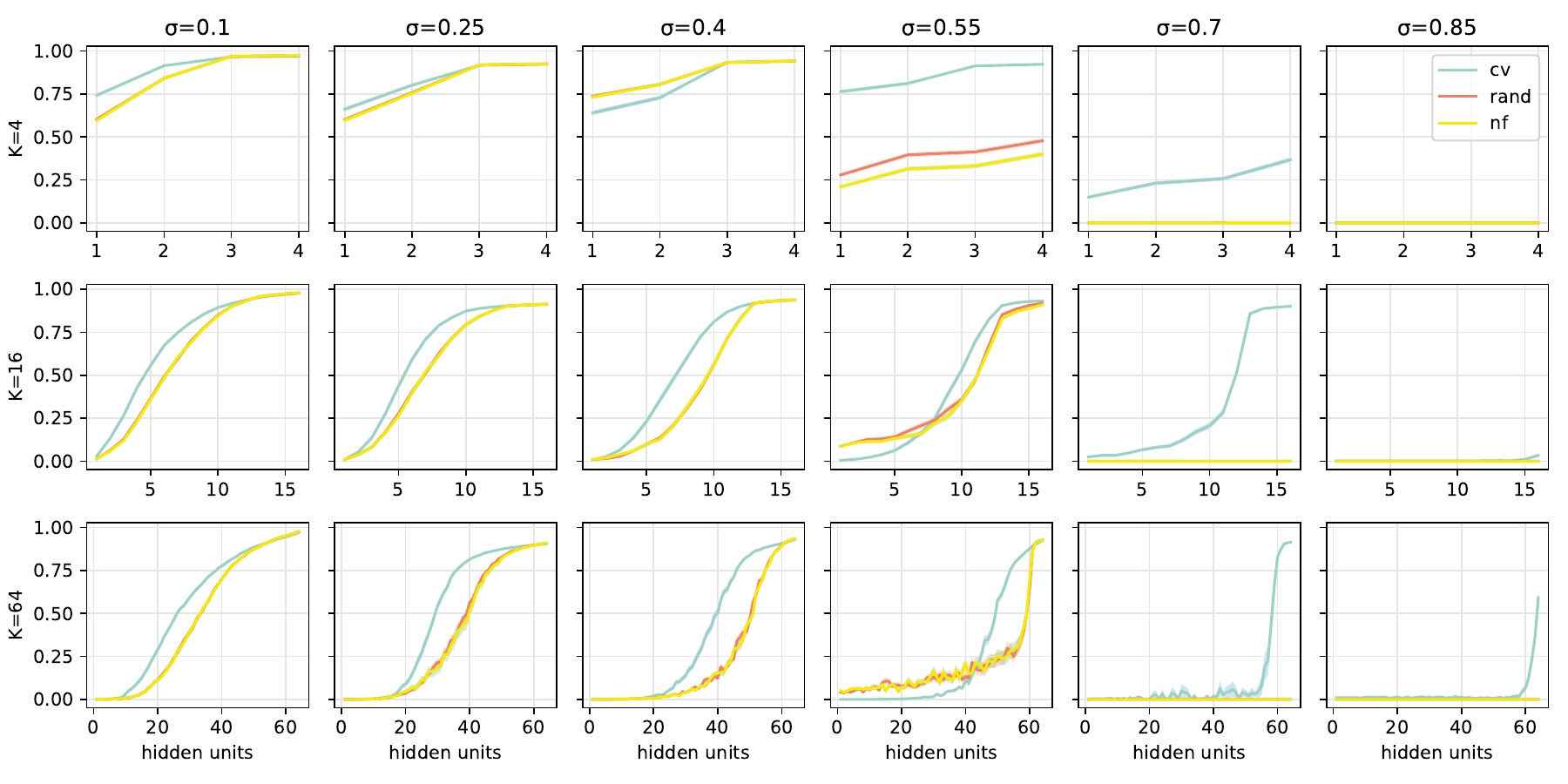}
    \end{center}
    \caption{\label{fig:ap-XOR-norm}
    \textbf{Norm on the relevant manifold}
    at different values of the noise, columns $\sigma\in\{0.1,0.25,0.4,0.55,0.7,0.85\}$, and degree of overparameterisation, rows $K\in\{4,16,64\}$. The individual plots show the norm of the relevant component of each hidden unit (ordered on the x-axis).
    }
\end{figure}

Finally in Fig.~\ref{fig:ap-XOR-norm}, we compute and show the norm on the relevant manifold, quantity whose relevance was already discussed in detail in the controlled setting of in Sec.~\ref{sec:curriculum}. In most of cases, curriculum helps to uncover the relevant manifold and improve performance. However this is not the case for $K=4$ and $\sigma=0.4$, resulting in the dip observed in Fig.~\ref{fig:ap-XOR-loss}.

\section{Additional details on the real-data simulations.}\label{app:real_data}

In the Fashion MNIST experiment, we consider a multi-layer perceptron with $L$ dense layers with intermediate widths $H$ --reported in Fig.\ref{fig:fig5}a with the label ``width''. We train the network for $100$ epochs, employing the
Adam optimiser with learning rate $\eta=0.001$ and batchsize $200$. In the \emph{curriculum} protocol, we split the training into two stages --$50$ epochs each-- and drop the learning rate by a factor $1/3$ between them. The same drop is also applied in the \emph{random order} protocol, except the dataset is not split in half according to difficulty in the two stages.

In the CIFAR10 experiment, we train a custom convolutional network with $5$ blocks containing a layer of $3\times3$ convolutions --with padding and stride $1$-- and a layer of MaxPool each. 
Given a filter factor $f$ --reported in Fig.\ref{fig:fig5}b with the label ``filters''-- the number of channels in each layer are respectively: $f$ in the first block, $2f$ in the second, $4f$ in the third, and $8f$ in the fourth and fifth blocks. Before the output, we also add to the network two dense layers with an intermediate width of $4096$. We train the network with Adam, $\eta=0.0003$, and batchsize $256$.
Notice that, in this experiment, the model is tested on samples with an intermediate difficulty $w=2\mathrm{pixels}$, between hard ($w=6\mathrm{pixels}$) and easy ($w=1\mathrm{pixels}$): this is to ensure that extracting information from the additional pixels of the centre of the image is partially useful also at test time --otherwise learning on the correct width of the distracting frame would be optimal. Each learning stage has $50$ epochs, but we employ early stopping --terminating training when zero error is achieved-- to avoid overfitting.

Notice that, to observe the reported phenomenology, one needs a setting where at least the following requirements are met:
\begin{itemize}
\item Both easy and hard patterns should be informative for the considered test task. 
\item Training on a pure easy dataset should lead to a better performance on the test task than training on a pure hard dataset.
\end{itemize}
For the first requirement, note that if the hard component of the dataset is too noisy it can become detrimental to the network and the curriculum phenomenology trivialises.
The second requirement might seem trivial at first. However, given the strong moment-matching capabilities of deep networks, it is possible to observe scenarios where training on samples with a noise level matched with that of the test set could lead to a better performance than training on clean patterns. Both conditions were checked to be verified in our two experiments.


\end{document}